\documentclass{article}
\usepackage{spconf,amsmath,graphicx,xcolor}

\usepackage[
subtle,
]{savetrees}

\usepackage{acronym}
\acrodef{ANN}{Artificial Neural Network}
\acrodef{ASR}{Automatic Speech Recognition}
\acrodef{ASG}{Auto Segmentation Criterion}
\acrodef{AUC}{area under the curve}
\acrodef{CCA}{Canonical Correlation Analysis}
\acrodef{CNN}{Convolutional Neural Network}
\acrodef{CTC}{Connectionist Temporal Classification}
\acrodef{DL}{Deep Learning}
\acrodef{EEG}{Electroencephalography}
\acrodef{ERP}{Event-Related Potential}
\acrodef{GAN}{Generative Adversarial Network}
\acrodef{Grad-CAM}{Gradient-weighted Class Activation Mapping}
\acrodef{GradNAP}{Gradient-adjusted Neuron Activation Profile}
\acrodef{G2P}{Grapheme-to-Phoneme}
\acrodef{CMUDict}{CMU Pronunciation Dictionary}
\acrodef{LSTM}{Long Short-Term Memory}
\acrodef{HMM}{Hidden Markov Model}
\acrodef{IPA}{International Pronunciation Alphabet}
\acrodef{LRP}{layer-wise relevance propagation}
\acrodef{MLP}{Multi-Layer Perceptron}
\acrodef{NAP}{Neuron Activation Profile}
\acrodef{NAvAI}{Normalized Averaging of Aligned Inputs}
\acrodef{NLP}{Natural Language Processing}
\acrodef{ReLU}{Rectified Linear Unit}
\acrodef{RNN}{Recurrrent Neural Network}
\acrodef{TTS}{text-to-speech}

\title{Gradient-Adjusted Neuron Activation Profiles for Comprehensive Introspection of Convolutional Speech Recognition Models}
\name{Andreas Krug,
	Sebastian Stober}

\address{Otto von Guericke University Magdeburg, Germany}
\begin{document}
%
\maketitle
\begin{abstract}
Deep Learning based Automatic Speech Recognition (ASR) models are very successful, but hard to interpret.
To gain better understanding of how Artificial Neural Networks (ANNs) accomplish their tasks, introspection methods have been proposed.
Adapting such techniques from computer vision to speech recognition is not straight-forward, because speech data is more complex and less interpretable than image data.
In this work, we introduce Gradient-adjusted Neuron Activation Profiles (GradNAPs) as means to interpret features and representations in Deep Neural Networks.
GradNAPs are characteristic responses of ANNs to particular groups of inputs, which incorporate the relevance of neurons for prediction.
We show how to utilize GradNAPs to gain insight about how data is processed in ANNs.
This includes different ways of visualizing features and clustering of GradNAPs to compare embeddings of different groups of inputs in any layer of a given network.
We demonstrate our proposed techniques using a fully-convolutional ASR model.
\end{abstract}

\section{Introduction}
\acp{ANN} have become a very popular tool for solving challenging tasks across various fields of application.
Performance gains are often achieved through increasing their complexity in terms of types of architectures or the number of neurons \cite{Szegedy2015}.
At the same time, larger computational models become harder to interpret \cite{Yosinski2015}.
This complicates detecting erroneous behavior and thus can be risky in critical applications.
Introspection techniques have been proposed to get insight into \acp{ANN} \cite{Zeiler2014,Selvaraju2016}.
However, these methods are often designed for certain applications or architectures.
In particular, many introspection techniques focus on images, as features are easy to interpret visually.

The complexity of \acp{ANN} is becoming closer to that of real brains.
Those have been studied in neuroscience for over 50 years.
Well-established methods in this field can be adapted to analyze \acp{ANN} \cite{Krug2017}.
Our work is inspired by a popular technique from neuroscience, the \ac{ERP}.
The \ac{ERP} technique is used for analyzing brain activity through \ac{EEG} \cite{Makeig2009}.
\acp{ERP} aim to measure brain activity for a particular fixed event (stimulus).
As the event is consistent across all \ac{EEG} measurements, aligning the data at this stimulus and averaging the signals yields event-specific information \cite{Luck2005}.
This way, in \acp{ERP}, variations in brain activity are averaged out.
We analyze \acp{ANN} similarly, but as their responses are deterministic, we average out data variations.
For example, in a speech recognition model, activity can be observed for a particular phoneme in audio recordings of different speakers and articulations.

In our work, we present \acp{GradNAP} as an \ac{ERP}-inspired analysis of \acp{ANN}, which combines and extends our previous work.
\acp{GradNAP} allow for a comprehensive analysis of features and representations in any layer of the network, as well as identification of neurons which respond to a particular group of inputs.
We demonstrate multiple ways to examine network responses of a fully-convolutional \ac{ASR} model using \acp{GradNAP}.

\begin{figure*}[h]
	\centering
	\includegraphics[width=0.93\linewidth]{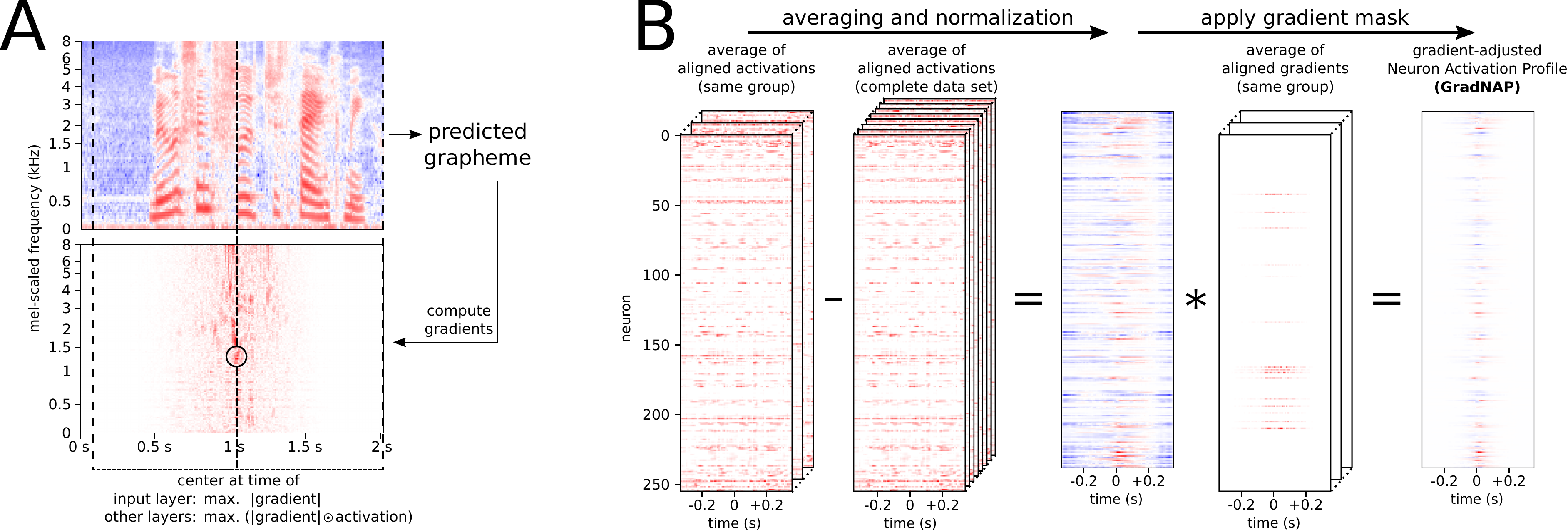}
	\caption{(A) Alignment procedure and (B) GradNAP computation from aligned activations for a layer.}
	\label{fig:GradNAP}
\end{figure*}

\section{Background}
\subsection{Convolutional speech recognition}
\acp{CNN} are not uncommon in \ac{ASR} \cite{Abdel2014}.
Here, we demonstrate our model using a simple, fully-convolutional architecture based on Wav2Letter \cite{Collobert2016}.
This architecture is useful for low-resource \ac{ASR} model training and transfer learning \cite{Kunze2017}.
Moreover, introspection methods from computer vision can easily be adapted to it \cite{Krug2018a}.
For comparability, we use a pre-trained model from our previous work \cite{Krug2018b}.
The 11-layer 1D-convolutional network predicts graphemes from spectrograms.
The model was trained on z-normalized spectrograms, which were scaled to 128 mel-frequency bins.
Whole sequence audio recordings from the LibriSpeech corpus \cite{Panayotov2015} were used as training data.
The acoustic model predicts sequences of graphemes, which are decoded by a \ac{CTC} beam search decoder.

\subsection{Model introspection for deep neural networks}
Introspection describes the process of analyzing or visualizing internal structures or processes of computational models.
This is of particular interest in \ac{DL} models, as these work as black-boxes \cite{Yosinski2015}.
Several introspection techniques have been proposed, mostly in the field of computer vision \cite{Zeiler2014,Selvaraju2016,Bach2015}.
A common way of explaining \acp{ANN} is to visualize learned features by optimizing the input to maximally activate certain neurons or sets of neurons \cite{Yosinski2015,Erhan2009,Mordvintsev2015}.
Optimal inputs do not always look natural.
This problem can be tackled by regularizing the optimization.
Another typical introspection strategy is to determine parts of the input, which are relevant for a certain prediction \cite{Zeiler2014, Selvaraju2016,Erhan2009}.
Those techniques visualize saliency maps on top of the input, which are easy to interpret.
However, as those methods work on single examples, it is hard to assess the model comprehensively.
Moreover, one has to choose such introspection techniques carefully, as some can be misleading \cite{Adebayo2018}.
More comprehensive insight into \acp{ANN} is provided by analyzing representations of different classes using the complete data set.
This can be done by training linear classifiers on intermediate representations \cite{Alain2016}, through \ac{CCA} of representations or by clustering class-specific neuron activations \cite{Nagamine2015}.
In speech, the latter type of analysis was conducted for \acp{MLP} for speech-to-phoneme prediction \cite{Nagamine2015,Nagamine2017} and for convolutional \ac{ASR} \cite{Krug2018a,Krug2018b}.

\section{Methods}
\subsection{Model \& data set}
For comparability with our previous work \cite{Krug2018b}, we use the same model and data set.
The architecture is based on Wav2Letter \cite{Collobert2016} and was trained on the LibriSpeech corpus \cite{Panayotov2015}.
This data set does not contain phoneme mappings.
Therefore, in our earlier work, we obtained them through a grapheme-to-phoneme translation model using an attention-based encoder-decoder architecture, trained on the \ac{CMUDict} \cite{lenzo2007cmu}.

\subsection{Gradient-adjusted Neuron Activation Profiles}
We introduce \acp{GradNAP} as a way to compute characteristic neuron responses of an \ac{ANN} to groups of inputs.
\acp{GradNAP} are an adaptation of the \ac{ERP} technique to \acp{ANN}.
Our method combines and extends two of our previously described introspection methods for \ac{ASR}: \ac{NAvAI} \cite{Krug2018a} and \acp{NAP} \cite{Krug2018b}.
We will first describe, how \ac{GradNAP} analysis differs from our previous work.
Afterwards, we explain our technique in detail.

As in our previous work, we adopt normalized averaging to obtain group-specific network responses.
To preserve more information than \acp{NAP}, we do not create time-independence by sorting on the time axis, but by sensitivity-based alignment like in \ac{NAvAI}.
Different to \ac{NAvAI}, we incorporate the activation strength into the alignment, as predictions can be highly sensitive to changes in inactive neurons.
On top of that, we mask out activations of low relevance for the prediction.
As these improvements utilize gradients, we call our method Gradient-adjusted \acp{NAP} (\acp{GradNAP}).

To properly apply an averaging approach, it is necessary that the different recordings are temporally aligned, similar to time-locked data in \ac{ERP} analysis.
To achieve this, we first center each layer's activations and the spectrogram frames at the time of highest importance for the prediction.
We refer to this step as ``alignment'' (Figure~\ref{fig:GradNAP}A).
We compute neuron activations and sensitivity values in every layer for each spectrogram frame.
Sensitivity is the gradient of a one-hot-vector for the predicted grapheme with respect to each layer's activations.
We identify importance for a prediction by strong activation with high absolute sensitivity value.
Hence, we center activations at time point $t$ of maximum  $(|\mathrm{gradient}|\odot\mathrm{activation})$.
As zeros in z-normalized spectrograms do not represent absence of the corresponding frequency, we center spectrogram inputs at time $t$ of maximum $|\mathrm{gradient}|$, 
The centering is implemented by cropping.
Equivalently, we center the gradients, so they remain aligned to the activations.

Figure~\ref{fig:GradNAP}B visualizes how to obtain a \ac{GradNAP} in a layer.
We average aligned activations and gradients over a group to obtain a group-specific profile.
As some neurons show baseline activations and some information are common to all inputs, we normalize activations by subtracting the average over the complete data set.
We do not normalize gradients this way, because zero-gradients would lose their meaning.
Instead, we scale them to a range of $[0,1]$.
Finally, we apply this gradient mask to the normalized averaged activations to obtain a \ac{GradNAP}.
Input layer \acp{GradNAP} are computed using spectrogram frames instead of activations.

\subsection{Visualization of group-specific features}
Here, we visualize \acp{GradNAP} as line plots, inspired by typical action potential plots of real neurons.
We compute group-responsiveness $r$ of neurons as the neuron-wise sum of absolute values in the corresponding \ac{GradNAP}.
A neuron $n$ can be positively or negatively responsive to a group.
Hence, we multiply $r$ by the sign of the sum of \ac{GradNAP} values (Equation~\ref{eq:responsiveness}).
\begin{equation}
	\label{eq:responsiveness}
	r_n=\mathrm{sign}\left(\sum_{i}{\mathrm{GradNAP}_n[i]}\right)\cdot\sum_{i}\left|\mathrm{GradNAP}_n[i]\right|
\end{equation}
We obtain the 5 most responsive neurons in terms of $|r_n|$ and compute a common optimal input. 
The optimization target is the joint pre-activation of those neurons at a single time point.
Pre-activations of positively and negatively responsive neurons are maximized and minimized, respectively (Equation~\ref{eq:loss}).
We did not optimize for each responsive neuron separately, as class-specificity is distributed across multiple neurons \cite{Morcos2018b}.
\begin{equation}
\label{eq:loss}
loss = - \sum_{n, r_n > 0}{preact_n} + \sum_{n, r_n<0}{preact_n}
\end{equation}
We apply L1 and L2 regularization on the input values, scaled depending on the receptive field size~$RF$ in the respective layer~$l$.
We scale L1 regularization by $15 / RF_l$ and L2 by $0.1 / RF_l$.
This dependence on $RF$ avoids that regularization becomes stronger for deeper layers.
Optimization is performed using Adam \cite{Kingma2014} with learning rate $0.05$ for 16~steps, initializing the input with random values from a normal distribution with $\mu=0$ and $\sigma=0.001$.

\subsection{Representation power of layers for different groups}
We apply hierarchical clustering with Euclidean distance and complete linkage to \acp{GradNAP} of graphemes and phonemes, like in our previous work \cite{Krug2018b}.
Differently, instead of using a fixed distance threshold for emergence of clusters, we apply different distance thresholds using percentiles 75\% to 95\% in steps of 5\%.
We evaluate the resulting clusterings by computing their Silhouette score \cite{Rousseeuw1987}.
This score is based on the difference between distances within clusters compared to the nearest other cluster.
Moreover, we average Silhouette scores over those 5 thresholds in each layer.
This ensures that the information is not specific to a particular parameter choice.
We compare those results between graphemes and phonemes.

\section{Results \& Discussion}
\subsection{Per-layer \acp{GradNAP}}
\begin{figure}[h]
	\centering
	\includegraphics[width=\linewidth]{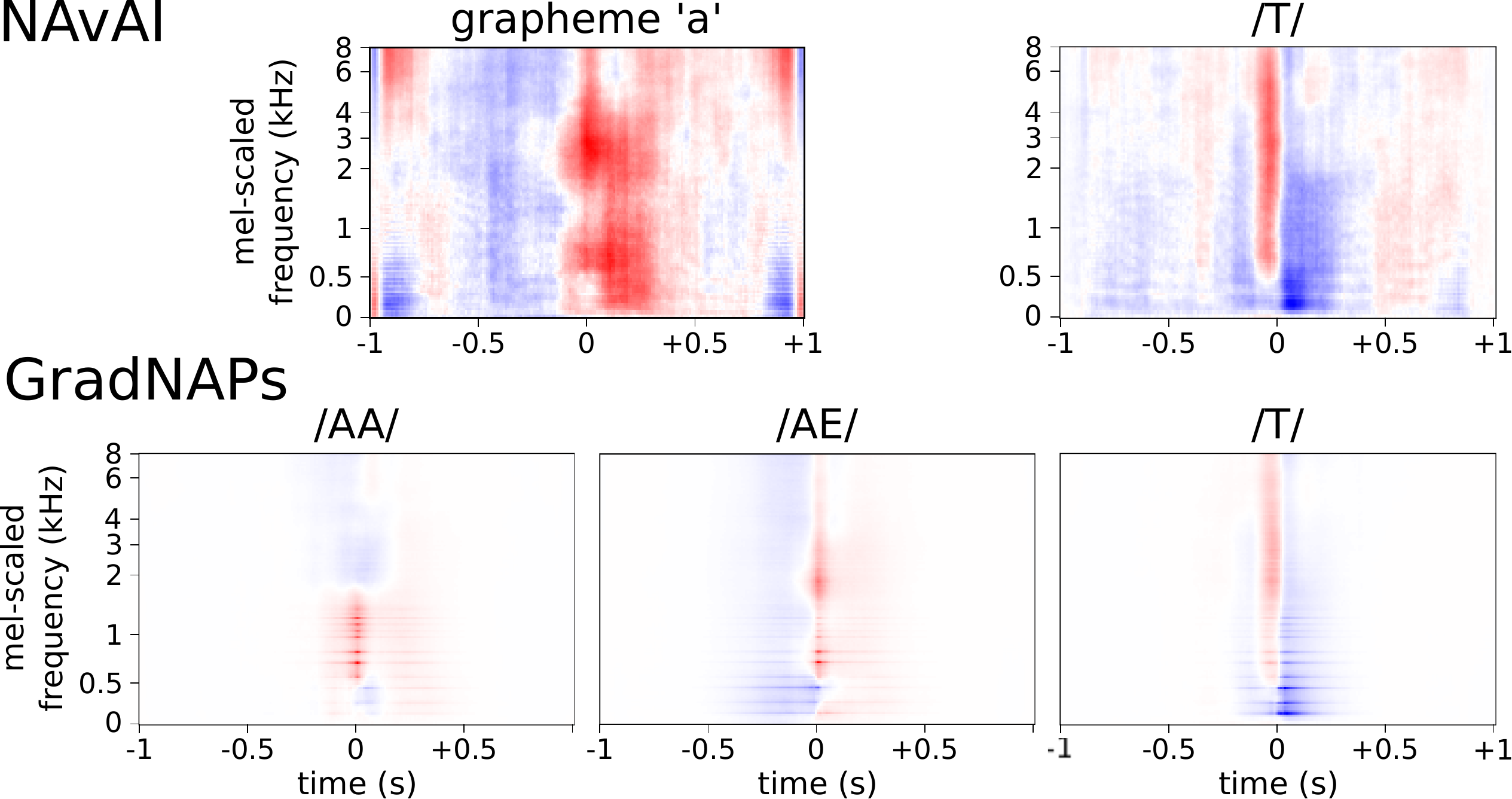}
	\caption{NAvAI patterns compared to input layer GradNAPs.}
	\label{fig:inputGradNAP}
\end{figure}
\acp{GradNAP} in the \textit{input layer} are an improvement of \ac{NAvAI} \cite{Krug2018a}.
Examples of \acp{GradNAP} in the input layer compared to exemplary \ac{NAvAI} results are shown in Figure~\ref{fig:inputGradNAP}.
\acp{GradNAP} in the input layer are directly interpretable.
They show how the intensity of frequencies differs from the average over the complete data set.
The gradient-based masking guarantees that the \ac{GradNAP} only shows regions, which are important for the prediction.
This advantage over \ac{NAvAI} is demonstrated for \texttt{/T/} in Figure~\ref{fig:inputGradNAP} (right).
While \ac{NAvAI} shows a pattern over the whole receptive field size, the corresponding \ac{GradNAP} also identified prediction-relevant parts of it.

We observe phoneme-typical patterns in input layer \acp{GradNAP} (Figure~\ref{fig:inputGradNAP} bottom).
Phonemes \texttt{/AA/} and \texttt{/AE/} share a high intensity formant at around 700~Hz.
A second formant is identified at around 1200~Hz and 1900~Hz for \texttt{/AA/} and \texttt{/AE/}, respectively.
The input pattern for \texttt{/T/} (right) shows a change of high to low intensities of all frequencies at the alignment time.
Those patterns match the expectation well.
However, identified formants are spreading a wider range of frequencies.
This is probably due to speaker variation.
The grapheme-specific \ac{NAvAI} result for \texttt{a} (as in \cite{Krug2018a}) is most similar to the input \ac{GradNAP} for \texttt{/AE/}.
This indicates that grapheme~\texttt{a} was pronounced as \texttt{/AE/} in the majority of the data.

In \textit{deeper layers}, neuron order does not have a meaning.
Therefore, corresponding \acp{GradNAP} cannot be interpreted by visual inspection.
An example can be seen in Figure~\ref{fig:GradNAP}B (rightmost).
Instead, we visualize features by optimizing inputs for the most responsive neurons.
Those results are shown in Section~\ref{sec:featurevis}.

In \textit{all layers}, we observed that \ac{GradNAP} values become smaller and drop to 0 the further away from the alignment time.
This indicates that the model did not use the complete receptive field for prediction.
Thus, compressing the model in terms of choosing smaller kernel sizes, fewer filters or layers is possible.

\subsection{Visualizing group-specific features}
\label{sec:featurevis}
\begin{figure}[h]
	\centering
	\includegraphics[width=\linewidth]{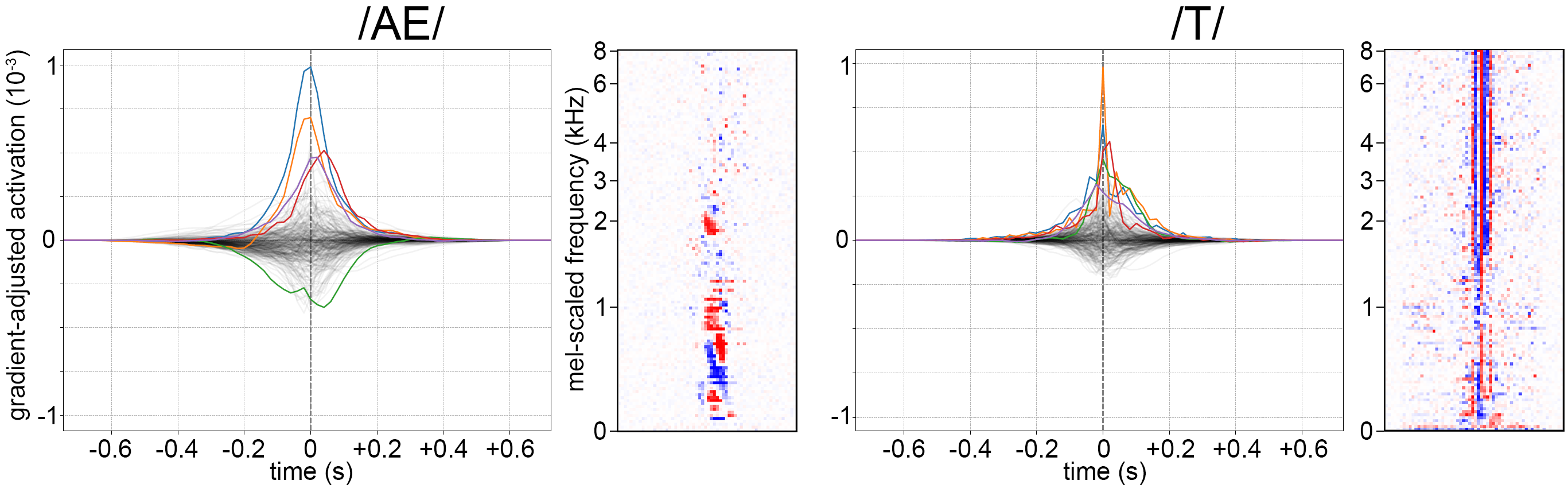}
	\caption{Neuron action potentials and feature visualization in the second layer for phonemes \texttt{/AE/} and \texttt{/T/}.}
	\label{fig:featureviz}
\end{figure}

Again inspired by neuroscience, we visualize \acp{GradNAP} as neuron action potentials.
Figure~\ref{fig:featureviz} shows \acp{GradNAP} of exemplary phonemes \texttt{/AE/} and \texttt{/T/} in the 2\textsuperscript{nd} layer as action potential plots.
The plots show phoneme-specific neuron activations for all neurons in the same layer superimposed.
The 5 most responsive neurons are highlighted with different colors (those do not represent the identity of the neuron).
We observed that neuron responses to both the vowel phoneme \texttt{/AE/} and the plosive \texttt{/T/} are close to the center.
This indicates that the network focuses on acoustic features of the phoneme, instead of correlating features in their context.
Next to each action potential plot in Figure~\ref{fig:featureviz}, the optimal input to the set of 5 most responsive neurons is shown.
The optimal input for neurons which are responsive to \texttt{/AE/} shows high intensities for frequencies around 700~Hz and 1900~Hz.
This corresponds to the \texttt{/AE/}-typical formants (also in agreement with Figure~\ref{fig:inputGradNAP}).
The visualized features for \texttt{/T/}-responsive neurons show several quick transitions from high to low intensities of most frequencies.
As no neuron peaks twice, it is likely that the multiple occurrence of the plosive pattern is related to detecting it in different contexts.
Because optimal inputs are not aligned, it is reasonable that features occur at more than one time point.
This also causes repetitive patterns in optimal inputs for deeper layers, which are distinguishable but not natural.
We omit them here, because unnatural feature visualization is not easily interpretable.
This problem could be tackled with stronger regularization, but could also lead to misleading interpretations.

\subsection{Analysis of grapheme and phoneme encoding}
\begin{figure}[h]
	\centering
	\includegraphics[width=\linewidth]{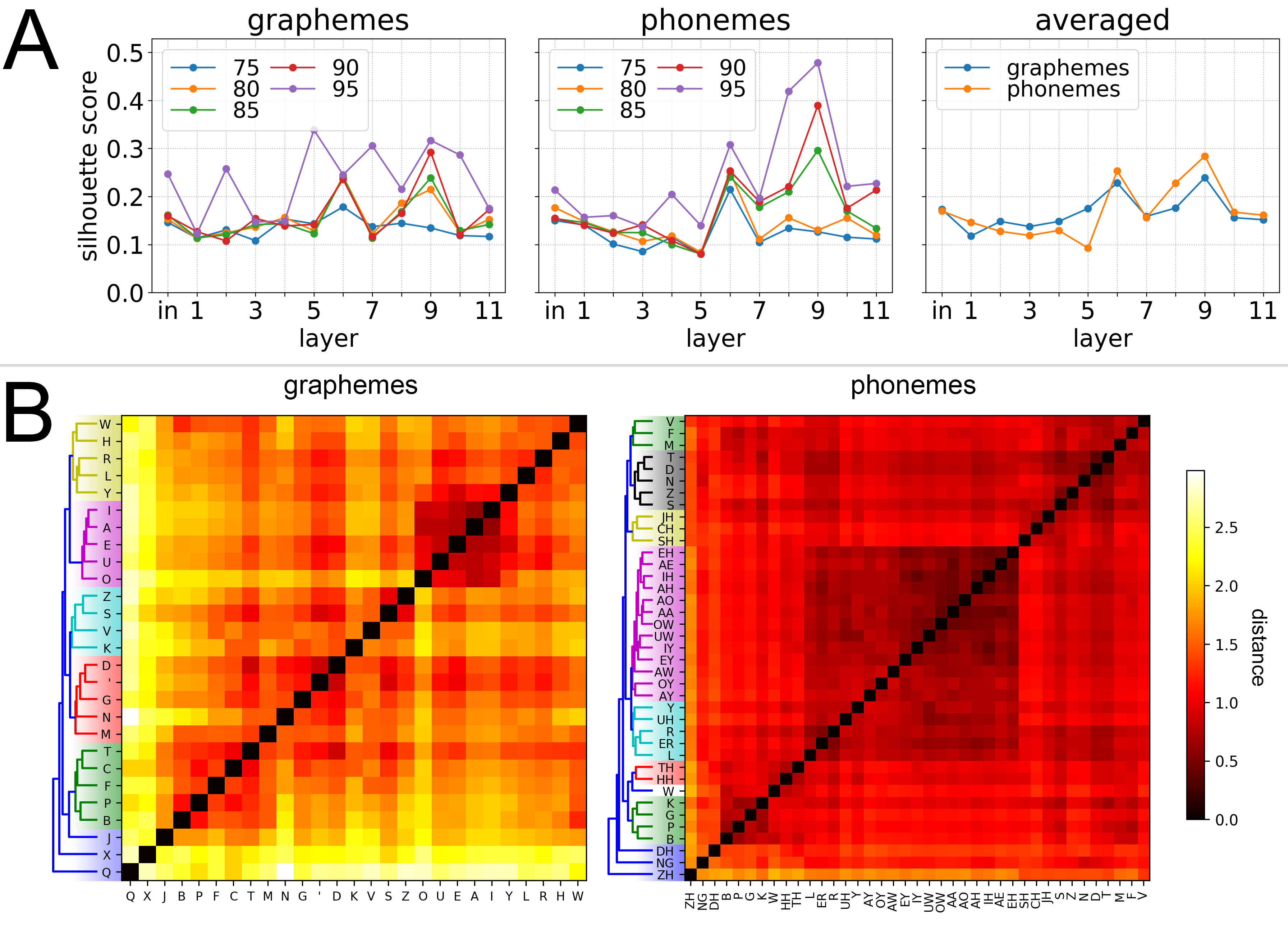}
	\caption{Silhouette scores at different distance thresholds (A) and 75\textsuperscript{th} percentile clustering of GradNAPs in layer~10 (B).}
	\label{fig:clustering}
\end{figure}

We analyze, which layers represent graphemes and phonemes best by clustering of \acp{GradNAP}.
In each layer, we compute Silhouette scores for cluster assignments using different distance thresholds.
Higher scores correspond to more distinct clusters, indicating better representation of the respective group.
Figure~\ref{fig:clustering} (top) shows Silhouette scores for graphemes (left), phonemes (center) and the averages over distance thresholds for both groupings (right).
Higher percentiles mostly lead to higher Silhouette scores.
This is reasonable, as we expect a hierarchy of similar phonemes rather than large clusters.
Surprisingly, representation quality does not consistently increase from lower to deeper layers.
Silhouette scores even decrease for phonemes from the input layer to the 5\textsuperscript{th} layer.
Deeper layers of the network show better clustering for phonemes than for graphemes over all distance thresholds.
The highest Silhouette score can be observed for phonemes in the 9\textsuperscript{th} layer.
However, the corresponding clusters are large and do not separate phonemic categories.
Layers 10 and 11 have a much larger number of neurons than the others.
This results in differently distributed distance matrices, which probably causes the drop of cluster quality from layer 9 to layer 10.

Silhouette scores indicate that in higher layers, phonemes are better represented than graphemes.
However, they are not suitable for detecting the exact layer, where clusters of meaningful phonemic categories emerge.
This indicates that phoneme similarity is not the strongest factor for distinguishing neuron responses.
Nevertheless, we observe similar phonemic categories in clusterings from the 10\textsuperscript{th} layer on, which is shown in Figure \ref{fig:clustering} (bottom).
In an earlier work, we performed clustering analysis of \acp{NAP} \cite{Krug2018b}.
We confirm the prior finding, that phonemic categories are represented well from the 10\textsuperscript{th} layer on and that the phoneme clustering is identifying more sub-categories.
However, the differences between grapheme and phoneme clustering are smaller than in our earlier work.
Most likely, this is an effect of gradient masking, which scales down a lot of prediction-irrelevant values.

\section{Conclusion}
\acp{GradNAP} are a promising tool to gain insight into \acp{ANN}.
We combined strengths of existing introspection techniques, extended them and applied more comprehensive analyses.
With our method, introspection is not limited to the predicted classes, but can be performed for any grouping of inputs.
Moreover, model introspection is possible for different parts of the network (inputs, any layer, subsets of neurons).
We presented per-layer clustering of \acp{GradNAP} for different groups and action potentials with feature visualization on the individual-neuron level.
Our method is generally applicable to any type of data and is not limited to \acp{CNN}.
If there are too many groups, the clustering overview can become cluttered.
This can be circumvented by choosing higher-level groups or only a subset of interest.
Future work will utilize our method to analyze the network during training.
This could shed light on when and how the network learns to detect features for particular groups.

\section{Acknowledgements}
This research has been funded by the Federal Ministry of Education and Research of Germany (BMBF)
and supported by the donation of a GeForce GTX Titan X graphics card from the NVIDIA Corporation.

\newpage

\bibliographystyle{IEEEbib}
\bibliography{strings,refs}

\end{document}